\documentclass[runningheads]{llncs}
\usepackage[T1]{fontenc}
\usepackage{subfig}
\usepackage{mathtools}
\usepackage{amsmath}
\usepackage{amssymb}

\usepackage{hyperref}

\usepackage{multirow}
\usepackage{graphicx}
\usepackage{booktabs} 
\usepackage[table]{xcolor}
\usepackage{cleveref}
\usepackage{color}

\usepackage{amsmath} 

\begin{document}
%
\title{Intraoperative Registration by Cross-Modal Inverse Neural Rendering}
%
%
\author{Maximilian Fehrentz\inst{1,2} \and
Mohammad Farid Azampour\inst{2} \and
Reuben Dorent\inst{1} \and
Hassan Rasheed\inst{1,2} \and
Colin Galvin\inst{1} \and
Alexandra Golby\inst{1} \and
William M. Wells\inst{1,3} \and
Sarah Frisken\inst{1} \and
Nassir Navab\inst{2} \and
Nazim Haouchine\inst{1}}

\authorrunning{Fehrentz et al.}

\institute{
Harvard Medical School, Brigham and Women's Hospital, Boston, MA, USA \and
Computer Aided Medical Procedures, TU Munich, Munich, Germany \and
Massachusetts Institute of Technology, Cambridge, MA, USA}

\maketitle              
\begin{abstract}
We present in this paper a novel approach for 3D/2D intraoperative registration during neurosurgery via cross-modal inverse neural rendering.
Our approach separates implicit neural representation into two components, handling anatomical structure preoperatively and appearance intraoperatively. 
This disentanglement is achieved by controlling a Neural Radiance Field's appearance with a multi-style hypernetwork. 
Once trained, the implicit neural representation serves as a differentiable rendering engine, which can be used to estimate the surgical camera pose by minimizing the dissimilarity between its rendered images and the target intraoperative image. 
We tested our method on retrospective patients' data from clinical cases, showing that our method outperforms state-of-the-art while meeting current clinical standards for registration. Code and additional resources can be found at \href{https://maxfehrentz.github.io/style-ngp/}{https://maxfehrentz.github.io/style-ngp/}.
\end{abstract}


\section{Introduction}

The use of surgical navigation techniques through patient-to-image registration has become a standard practice in neurosurgery~\cite{marcus2015comparative}.
It allows neurosurgeons to visualize preoperative imaging during the operation, enabling them to achieve a maximal safe tumor resection that is highly correlated with patients' chances of survival~\cite{Sanai} and has been shown to reduce risks of postoperative neurological deficits~\cite{Gonzalez-Darder2019}.
In this paper, we address patient-to-image registration in neurosurgery as a 6-degrees-of-freedom (DoF) pose estimation problem. 
This process involves aligning preoperative Magnetic Resonance (MR) images with intraoperative surgical views of the brain surface revealed after a craniotomy and acquired using a camera.
Different from previous approaches~\cite{Kuhnt_2012,Pereira,Filipe,Nakajima,Mohammadi,JI20141169}, our proposed method 
relies solely on imaging already available in the operating room, eliminating the need for cumbersome and time-consuming imaging acquisitions or optical tracking systems. 

In most cases, tackling 3D/2D registration in surgery involves bridging the preoperative to intraoperative modality gap and resolving 3D to 2D projection ambiguities.
3D shape reconstruction of surgical scenes has been proposed as a way to tackle both issues at once~\cite{Stoyanov2012}. 
They provide a modality-agnostic 3D surface representation of the surgical scenes (3D point clouds or meshes) and re-cast the 3D/2D registration as a 3D-3D point set registration problem where robust methods exist~\cite{Stoyanov2012}.
Other approaches rely on landmark-based matching~\cite{Luo,haouchine2023learning,koo2022automatic}, where anatomical landmarks, extracted from both modalities, can act as image abstractions. 
Iterative registration methods can then be applied with a closed-form when landmarks are paired which often involve surgical pointers, tracking systems, or \textit{in-vivo} markers~\cite{Luo,haouchine2023learning,koo2022automatic}. 

During neurosurgery, the brain surface is revealed and viewed using a surgical camera (microscope), and although the field of view is limited w.r.t other organs where a larger part of the organ is visible, it has the advantage of having visible vessels at the cortical level.
These vessels have been used as salient sources of information to drive 3D/2D registration.
In~\cite{Haouchine2020}, segmentations of cortical vessels are used to drive a 3D/2D non-rigid registration. 
Instead of using segmentations, the authors in~\cite{Luo} proposed to manually trace vessels at the brain surface that match preoperative scans.
This method, however, involves a tracked pointer. 
Other methods proposed to pre-compute the set of plausible transformations preoperatively, using atlas-based approaches~\cite{Sun} or by learning to estimate poses \cite{Haouchine2022} and appearances~\cite{haouchine2023learning}. 
However, they are trained in a patient-specific manner on a pre-defined set of transformations and may fail with out-of-distribution transformations. 
Outside of surgical applications, the field of computer vision has seen the emergence of differentiable rendering~\cite{kato2018neural} and Neural Radiance Fields (NeRFs)~\cite{nerf} making approaches for 6-DoF pose estimation more robust.
For instance, methods such as iNeRF~\cite{inerf} and Parallel Inversion~\cite{parellel_inversion} showed that implicit neural representations outperform conventional regression-based methods.
These methods, however, are not designed for multimodal registration, where the appearance of the learned representations differs from the one of the target images, as is the case for intraoperative registration. Although NeRFs have recently been used for 3D reconstruction of endoscopic scenes demonstrating remarkable performances~\cite{wang2022neural,zha2023endosurf}, their utilization for multimodal registration remains unexplored.

\paragraph{\textbf{Contribution}} In this work, we propose a novel 3D/2D registration approach for single-view neurosurgical registration using implicit neural representations. 
We introduce a new formulation that separates NeRFs into structural and appearance representation, where the anatomical structure is learned preoperatively and appearance is adapted intraoperatively. 
This is achieved by training a hypernetwork that controls the appearance of the NeRF as proposed in \cite{chiang2022stylizing} while leaving its learned representation of the anatomy untouched. Given a single intraoperative image, the hypernetwork crosses the modality gap and enables the NeRF to solve the 6-DoF pose estimation problem.
Experiments on synthetic and real data demonstrate the effectiveness of the proposed approach, outperforming the state-of-the-art methods.

\begin{figure}[ht!]
     \begin{center}
        \includegraphics[width=1\linewidth]{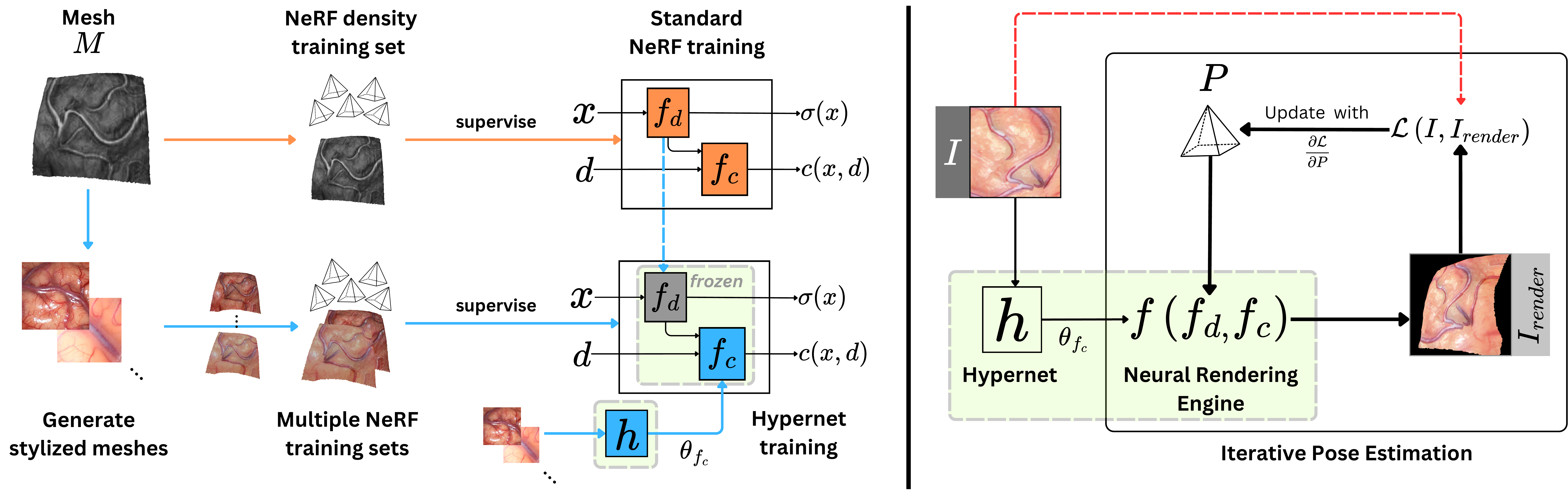}
    \end{center}
    \caption{Left (preoperative): We use a NeRF to first learn density/anatomy (orange) from a mesh $\mathbf{M}$ extracted from an MR scan, then learn single-shot style adaptation (blue) through a hypernetwork $h$ while freezing the rest of the NeRF, keeping the previously learned density $f_d$ fixed. Right (intraoperative): Iterative pose estimation on target $\mathbf{I}$. The trained NeRF and hypernet (green highlights) are used as style-conditioned neural rendering engine using ray marching, with $f$ adapted to the appearance of the intraoperative registration target $\mathbf{I}$ through the hypernetwork $h$.}
    \label{fig:overview}
\end{figure} 

\section{Methods}

\subsection{Problem Formulation \& Overview}

Given a preoperative surface mesh $\mathbf{M}$ of the craniotomy area and an intraoperative image $\mathbf{I}$ obtained from a surgical microscope, we seek to determine the pose $\mathbf{P} \in SE(3)$. 
This pose minimizes a loss function $\mathcal{L}(\mathbf{P}| \mathbf{I}, \mathbf{M})$, quantifying the discrepancy between the observed intraoperative image and the preoperative mesh $\mathbf{M}$ when positioned and oriented according to the pose $\mathbf{P}$.

We approach the problem as an optimization in 2D image space. Our method minimizes the loss between $\mathbf{I}$ and images rendered from a continuous and differentiable neural representation $f$ of the craniotomy area.
The density of $f$ is learned preoperatively from $\mathbf{M}$ and its appearance is controlled intraoperatively by $\mathbf{I}$. This approach effectively addresses the 3D-to-2D registration problem by using $f$ to render 2D images based on the anatomy learned from $\mathbf{M}$. It also bridges the modality gap by incorporating intraoperative appearance conditioning on $\mathbf{I}$.
We therefore reformulate the optimization problem with our neural representation $f$ and an image-based loss $\mathcal{L}$, yielding:
\begin{equation}
  \hat{\mathbf{P}} = \underset{\mathbf{P} \in SE(3)}{\mathrm{argmin}} \; \mathcal{L}(\mathbf{I}, f(\mathbf{P} | \mathbf{I}, \mathbf{M}))
\label{eq:image_problem}
\end{equation}

Given that $f$ is differentiable with respect to $\mathbf{P}$, we can find $\hat{\mathbf{P}}$ through iteratively rendering $f(\mathbf{P} | \mathbf{I}, \mathbf{M})$ and optimizing $\mathbf{P}$.

\subsection{Bridging the Domain Gap via Hypernetwork Multi-Style NeRF}
\subsubsection{NeRF Background.}
A Neural Radiance Field (NeRF)~\cite{nerf} represents a continuous neural representation of a 3D scene. For our method, we follow Instant-NGP~\cite{instant_ngp}, a fast NeRF based on hash-grid encodings. The NeRF consists of a density (structure) and a color (appearance) component, evaluating a single point and viewing direction in space. They are mathematically described by the following equations:
\begin{equation}
\sigma(\mathbf{x}), z(\mathbf{x}) = f_{\text{d}}(\mathbf{x}; \theta_{f_d}),
\label{eq:nerf1}
\end{equation}
where $\sigma(\mathbf{x})$ is the density and $z(\mathbf{x})$ is an intermediate representation used as input for the RGB component, and
\begin{equation}
c(\mathbf{x}, \mathbf{d}) = f_{\text{c}}(z(\mathbf{x}), \mathbf{d}; \theta_{f_c}),
\label{eq:nerf2}
\end{equation}
where the color $c$ depends on $z(\mathbf{x})$ and the viewing direction $\mathbf{d}$. Both components consist of a Multilayer Perceptron (MLP), resulting in two sets of parameters, $\theta_{f_d}$ for density and $\theta_{f_c}$ for color. In practice, each component has a parametrized input encoding function, whose parameters are omitted here for clarity.

The continuous neural representation allows for rendering images by marching rays, evaluating the NeRF along each ray's path, and accumulating RGB values according to densities. For our method, NeRF serves as a neural renderer encoding our 3D mesh $\mathbf{M}$. This differs from traditional mesh representations since it is fully differentiable and has learnable disentangled components for structure $f_{d}$ and appearance $f_{c}$. Both aspects are key to our method, the former for iterative pose estimation, the latter to bridge the domain gap to $\mathbf{I}$.

\subsubsection{Training the Hypernetwork Multi-Style NeRF.}
First, we train a NeRF following Eq.~\ref{eq:nerf1} and Eq.~\ref{eq:nerf2} on a dataset from the preoperative MR-derived surface mesh $\mathbf{M}$. 
The objective is to capture the anatomical structure with high fidelity, resulting in a NeRF that faithfully replicates the brain surface from the MRI.

We introduce a hypernetwork that enables adapting to the intra-operative image appearance in real-time using only one single image. 
The hypernetwork, that we denote $h$, takes the form of a multi-head MLP and is trained to set the parameters $\theta_{f_c}$ of $f_c$ based on the appearance of $\mathbf{I}$, leaving the structure, encoded in $f_{d}$, untouched.
To train $h$, we use Neural Style Transfer (NST) to generate multiple training datasets by stylizing $\mathbf{M}$ using a set $N$ of style images $\{\mathbf{J}_i\}_{i=1}^N$ (obtained from other surgeries).
We combine two NST approaches, WCT\textsuperscript{2}~\cite{wct2} for its strong preservation of semantic, and STROTSS~\cite{strotss} for a more photorealistic style.
This changes the NeRF formulation to the following.
\begin{equation}
    \sigma(\mathbf{x}), z(\mathbf{x}) = f_{d}(\mathbf{x}; \hat{\theta}_{f_d}),
\label{eq:density_hyper}
\end{equation}
\begin{equation}
    c^{i}(\mathbf{x}, \mathbf{d}) = f_{c}^{i}(z(\mathbf{x}), \mathbf{d}; h(\mathbf{J}_i; \theta_h)),
\label{eq:rgb_hyper}
\end{equation}
where only $h$ is trained to learn $\theta_h$, while $\hat{\theta}_{f_d}$ has already been learned in the previous step and remains fixed. The whole pipeline is shown in Fig. \ref{fig:overview}.
In practice, $h$ uses a binned histogram of $\mathbf{J}_i$ instead of the whole image, since the structural information is already provided by $f_{d}$, allowing for simple low-dimensional color-based features. 
All training is done preoperatively, implemented in the Nerfstudio environment \cite{nerfstudio} following their implementation of Instant-NGP.

\subsection{Intraoperative Registration}
During surgery, assuming that camera intrinsics are known, we use $f_c$ and $f_d$ for a rendering engine $f$ that renders images via ray marching. Thus we can reformulate the  optimization problem in Eq.~\ref{eq:image_problem} as:
\begin{equation}
    \hat{\mathbf{P}} = \underset{\mathbf{P} \in SE(3)}{\mathrm{argmin}} \; \mathcal{L}\big(\mathbf{I}, f(\mathbf{P}; \hat{\theta}_{f_d}, h(\mathbf{I}; \hat{\theta}_h))\big)  
\label{eq:problem_reformulated}
\end{equation}
where $\mathbf{\hat{P}}$ represents the optimal camera pose, and $\mathcal{L}$ denotes the loss function comparing the intraoperative image $\mathbf{I}$ with the rendered image.
The intraoperative image $\mathbf{I}$ is used twice, as the target image, and as the input to $h$ to condition $f$ such that the rendered image approximates the appearance of $\mathbf{I}$.
This is key in our method since it allows us to bridge the modality gap and express $\mathcal{L}$ as a conventional, mono-modal, RGB image loss that takes the form of a relative $L_2$ loss \cite{parellel_inversion}.

The differentiable nature of $f$ allows computing $\frac{\partial \mathcal{L}}{\partial \mathbf{P}}$, thereby enabling iterative pose refinement via gradient descent. 
Multiple recent works propose efficient methods to solve Eq.~\ref{eq:problem_reformulated} \cite{parellel_inversion,inerf}.
We choose Parallel Inversion~\cite{parellel_inversion} that achieves state-of-art pose accuracy. 


\section{Experiments and Results}

\subsubsection{Datasets.}
We test our method on 5 clinical cases, each with its preoperative T1 MRI scan and corresponding surgical microscope image, except for case 5 with only a T1 MRI scan. 
We used T1 MRI from the ReMIND dataset \cite{juvekar2024remind}.
To stylize meshes, we rely on a small dataset of $N=15$ surgical microscope images from different cases, including the 4 corresponding to our clinical test cases that have a surgical microscope image. 
We build one dataset per case. 
To train the NeRF, we generate 100 images per style with their respective camera poses. 
For each case, this results in 1500 images that show the stylized brain surface mesh in 15 styles with 100 images and poses for each style. 
Additionally for case 5, for each of the 4 styles that correspond to the other clinical cases, we generate 50 random images and poses as test targets for registration.

\captionsetup[subfigure]{labelformat=empty}
\begin{figure}
\subfloat{\includegraphics[width=0.258\linewidth]{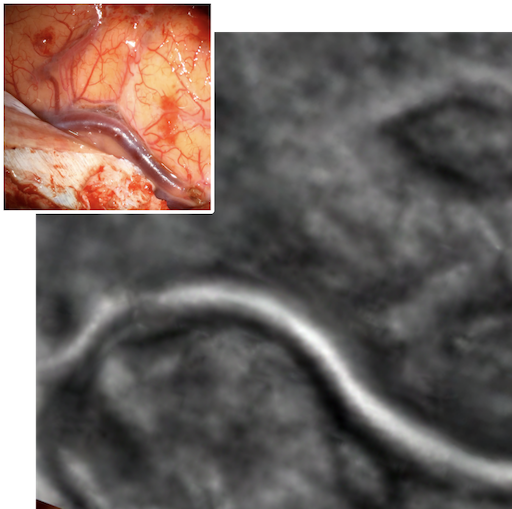}}
\hfill
\subfloat{\includegraphics[width=0.24\linewidth]{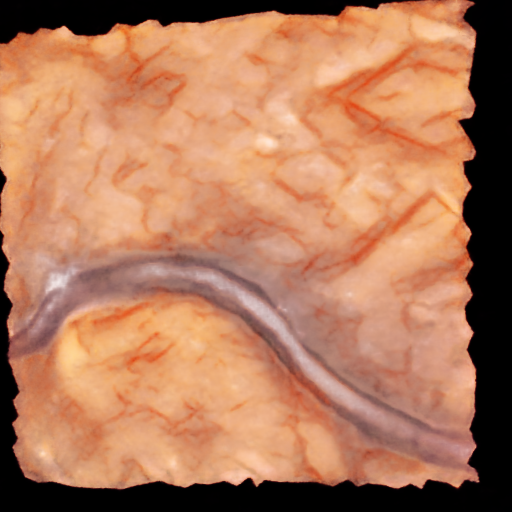}}
\hfill
\subfloat{\includegraphics[width=0.24\linewidth]{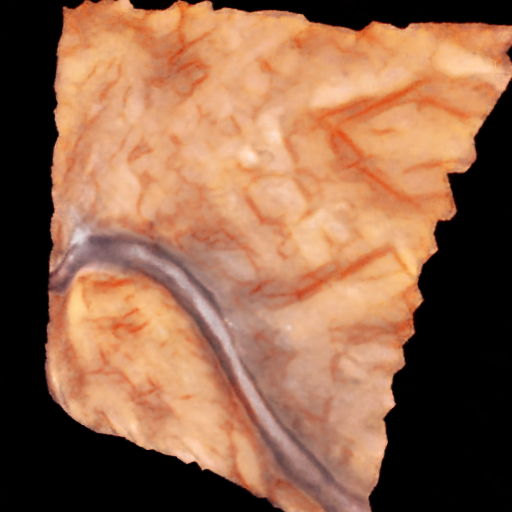}}
\hfill
\subfloat{\includegraphics[width=0.24\linewidth]{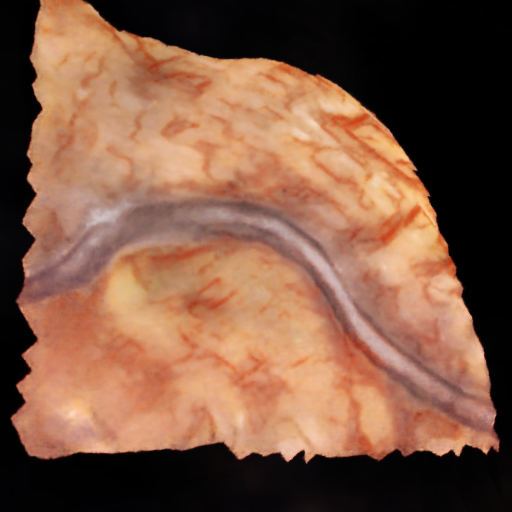}}\\
\caption{An example of synthesis from 3 different poses on one of the clinical cases. First: image obtained from the MRI with surgical microscope image $\mathbf{I}$. Remaining images: synthesis with $f$, style inferred by hypernetwork $h$ on $\mathbf{I}$.}
\label{fig:synthesis}
\end{figure}

\subsubsection{View Synthesis.}
To demonstrate that our hypernetwork produces plausible appearances, we synthesize 3 views from different poses on one of the clinical cases, as illustrated in Fig.~\ref{fig:synthesis}.
These syntheses are obtained by training on the dataset for this case while omitting its style. 
Our method shows qualitatively photorealistic results that respect the anatomy of the case while being similar in appearance.

\subsubsection{Pose Estimation on Synthetic Targets.}
\begin{figure}
\subfloat[Style variance]{\includegraphics[width=0.22\linewidth]{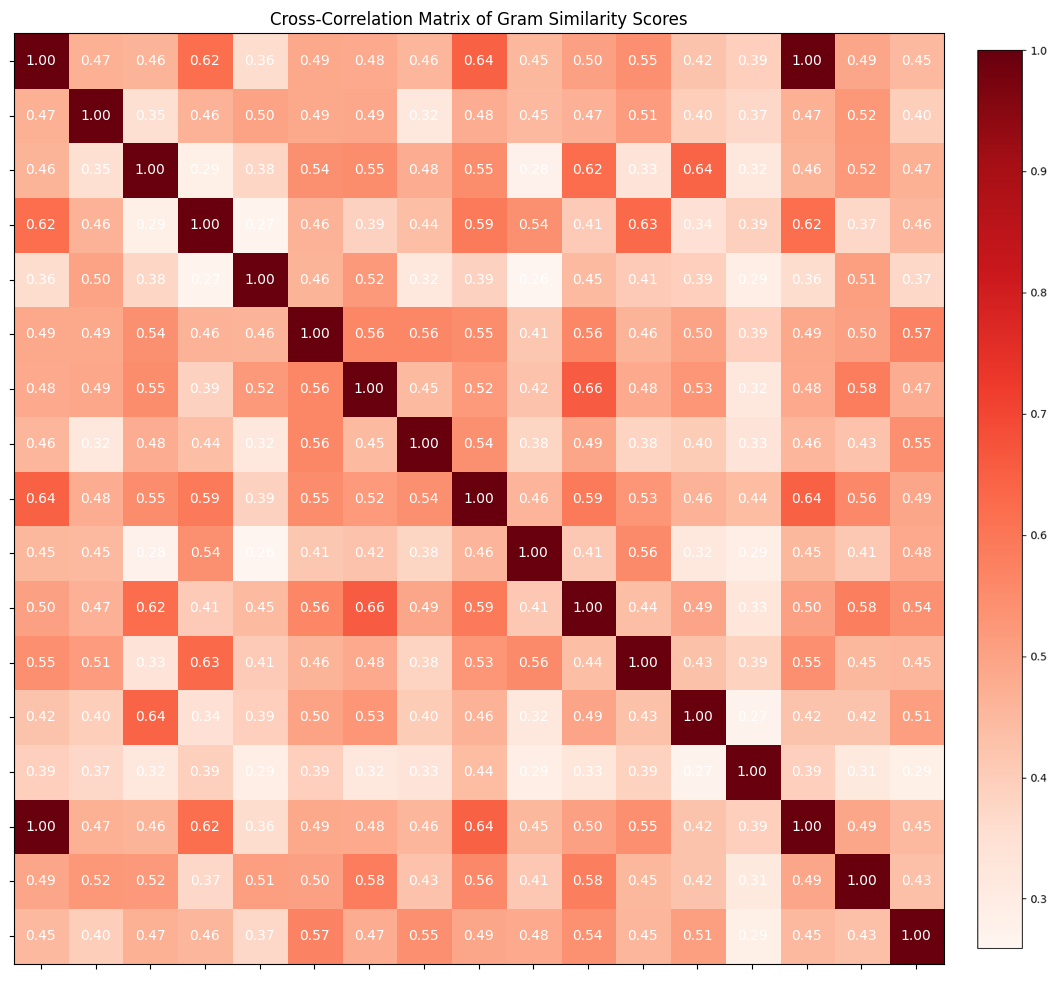}}
\hfill
\subfloat[Pose sampling]{\includegraphics[width=0.23\linewidth]{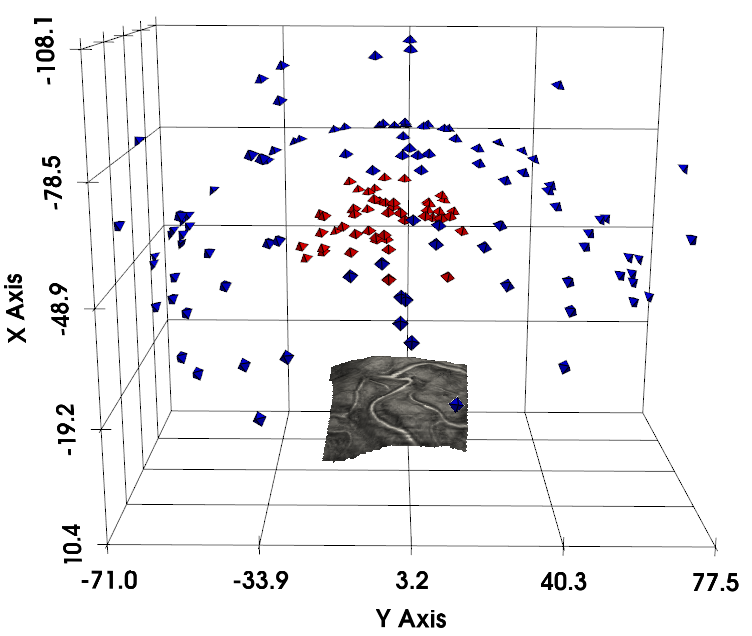}}
\hfill
\subfloat[Rotations (deg)]{\includegraphics[width=0.27\linewidth]{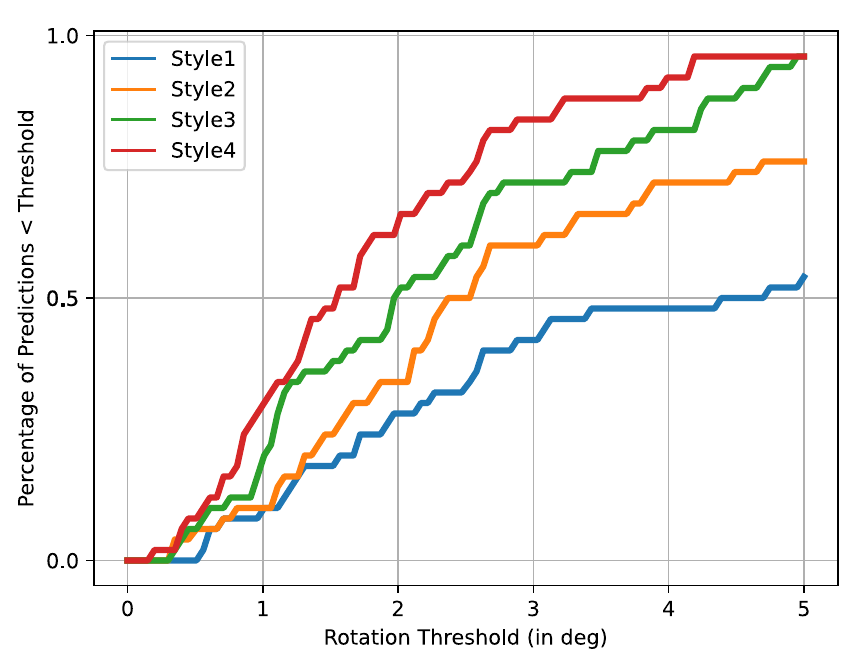}}
\hfill
\subfloat[Translations (mm)]{\includegraphics[width=0.26\linewidth]{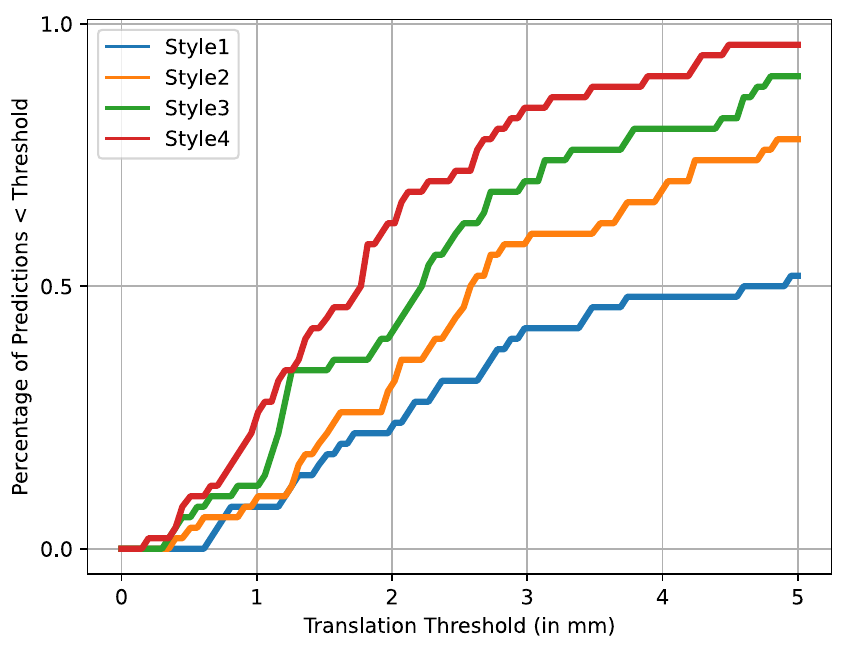}}
\hfill
\caption{Evaluation on synthetic targets (from left to right): cross-correlation matrix of Gram-similarity score of all styles showing pairwise style similarity and dissimilarity; pose distribution (blue: training set, red: test set); and accuracy-threshold curves for rotation and translation.}
\label{fig:acc_thres}
\end{figure}

To evaluate our method on a larger number of registration targets, we use case 5 and generate 50 targets for 4 different styles (corresponding to the clinical test cases), yielding 200 targets in total. 
The hypernetwork is trained on the remaining 11 styles. 
For each style, the hypernetwork conducts a singular inference to determine the style-specific appearance, which is then utilized in the pose estimation for all 50 targets within that style. 
The results are shown as accuracy-threshold curves (Fig. \ref{fig:acc_thres}), which indicate the proportion of predicted poses that fall within a given error threshold. 
Pose estimation on a $256 \times 256$ image takes 90 seconds on an NVIDIA GeForce RTX 2080 Ti. 

\captionsetup[subfigure]{labelformat=empty}

Fig. \ref{fig:acc_thres} shows that rotation and translation errors vary depending on the style of the targets.
While $96\%$ of poses estimated for Style 3 and Style 4 targets have a rotation error below 5$^{\circ}$, it is $76\%$ for targets of Style 2. 
Similarly, $96\%$ of poses estimated for Style 4 and $90\%$ of poses estimated for Style 3 have a translation error below 5\,mm, whereas the pose estimations on targets of Style 2 reach $78\%$. 
Style 1 on the other hand has $52\%$ of poses with translation error below 5\,mm and $54\%$ of poses with rotation error below 5$^{\circ}$. 
This can be explained by the small number of style images in the hypernetwork dataset ($N=11$).

\subsubsection{Comparison with Baseline and State-of-the-art.}
We evaluate our method \textbf{Ours} against a baseline and the state-of-the-art method. 
The baseline is our NeRF trained exclusively on images and poses from the MRI visualized as volume rendering, referred to as \textbf{MR-NeRF}. 
The state-of-the-art method is a style-invariant regressor \cite{haouchine2023learning}. 
We used case 5 data with 11 styles of 100 images and poses each, and evaluated on a test set of 4 styles of 50 images and ground-truth poses each. 
We trained and tested two versions of the regressor, \textbf{Reg.} with the described data also used for \textbf{MR-NeRF} and \textbf{Ours}, and \textbf{Reg. ID}, with a modified training and test split with poses in-distribution, following the method guidelines.
The pose distribution is visualized in Fig~\ref{fig:acc_thres}, where red poses correspond to the test set and blue ones to the training set.
The style distribution is also shown in Fig~\ref{fig:acc_thres} with the cross-correlation matrix of Gram-similarity scores, a common similarity measure used in Neural Style Transfer. 

For all methods, we report the Average Translation Error (ATE) and Average Rotation Error (ART) in Tab.~\ref{tab:comparative_registration_error}.
We also define outliers as poses with a rotation error larger than 20$^{\circ}$. 
Outliers are excluded from ATE and ART and reported separately as a percentage of the total number of test poses. 
For \textbf{Ours}, we additionally report results on all styles individually.

\def\arraystretch{0.8}
\rowcolors{2}{}{gray!10}
\begin{table}[t!]
\centering
\caption{Comparative Registration Error}
\begin{tabular}{l|c|c|c|c!{\vline width 1.5pt}c|c|c|c}
\toprule
Metrics / Methods  &  Style1 & Style2 & Style3 & Style4  &  \textbf{Ours}  &  MR-NeRF  &  Reg.  &  Reg. ID \\ 
\midrule                    
ATE (mm) & 4.17 & 3.24 & 2.56 & 1.96 & \textbf{3.12} & 10.12  &  13.31 & 6.29              \\ 
ART (deg)  & 4.53  & 3.25 & 2.41 & 1.84 & \textbf{3.01} & 11.30  &  8.78 & 5.70               \\ 
Outliers (\%)  & 22    & 6    & 0    & 0 & 7   &  46.50   &  58 & 4.52  \\ 
\bottomrule
\end{tabular}
\label{tab:comparative_registration_error}
\end{table}

Our method outperforms both baseline and state-of-the-art and achieves ATE and ART of 3.12mm and 3.01${^\circ}$ that meet clinical needs \cite{Frisken}. 
We also achieve style-invariance for Style 2, Style 3, and Style 4. 
However, we do not achieve style-invariance for Style 1 with a high number of outliers, in line with the Accuracy-threshold curve experiment. 
Given that the underlying anatomy and target poses are the same across all styles, this indicates that the appearance approximated by our hypernetwork is not sufficiently similar enough to Style 1 to allow for a robust pose estimation with gradient descent. 
The \textbf{MR-NeRF} is not able to cross the modality gap to the target images while \textbf{Reg.} does not generalize to out-of-distribution target poses. 
Only when withholding and evaluating part of the more homogeneous training set does the performance of the regressor improve with \textbf{Reg. ID} as shown in Tab.~\ref{tab:comparative_registration_error}.

\subsubsection{Tests on Clinical Cases.}
\begin{figure}
\subfloat{\includegraphics[width=0.16\linewidth]{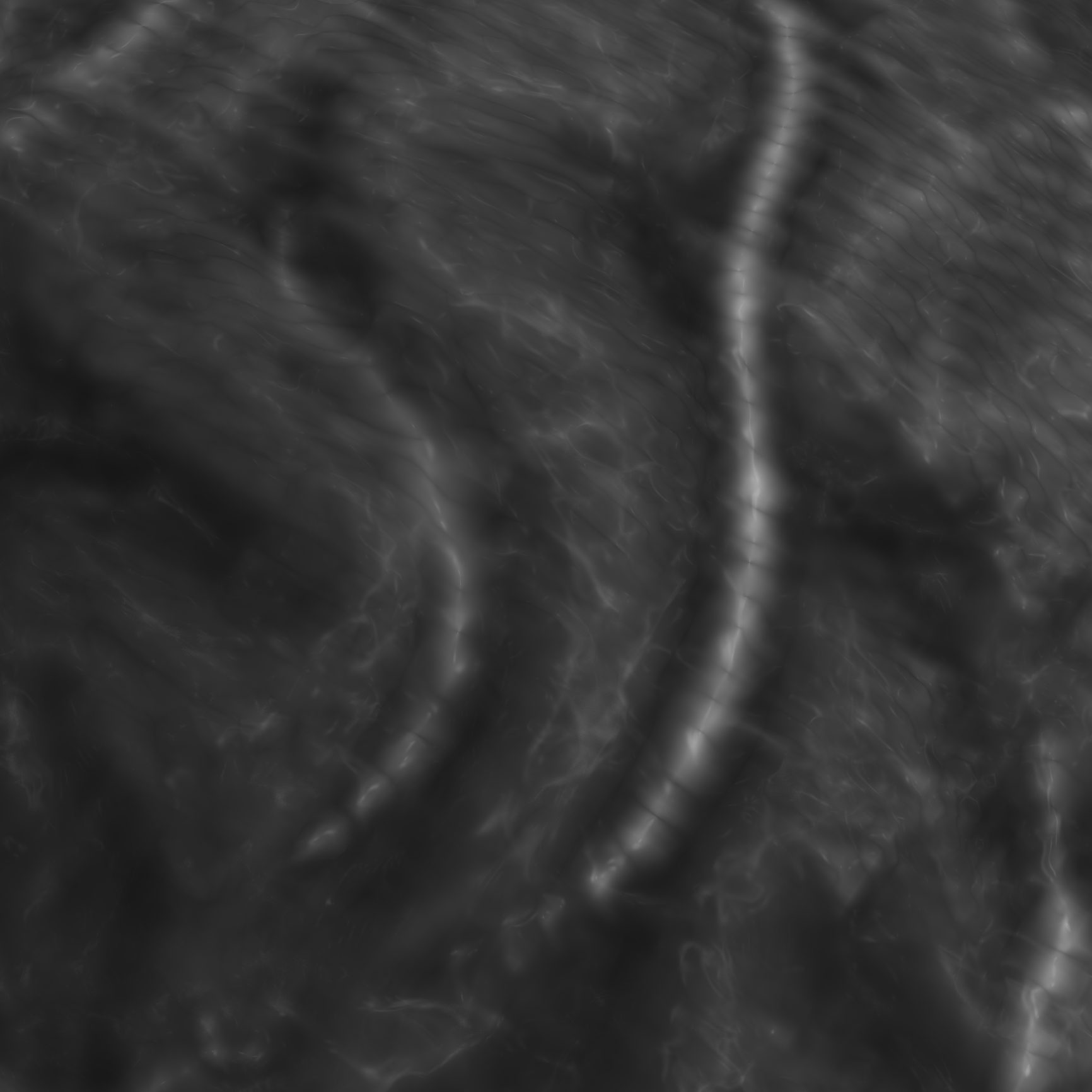}}
\hfill
\subfloat{\includegraphics[width=0.16\linewidth]{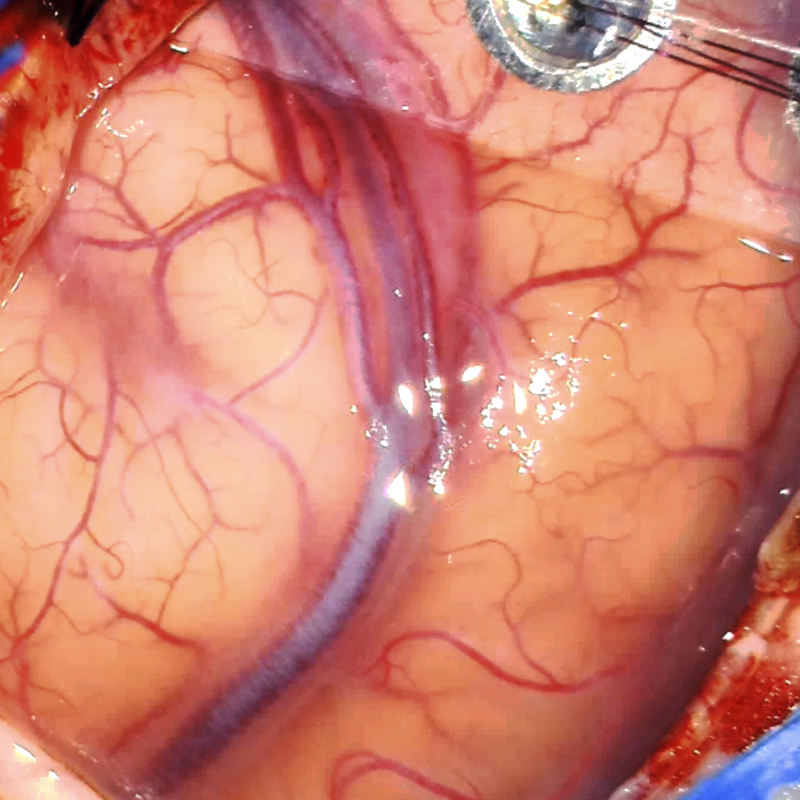}}
\hfill
\subfloat{\includegraphics[width=0.16\linewidth]{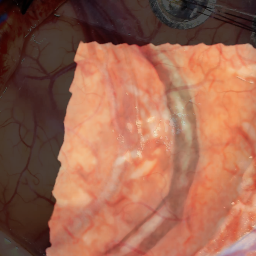}}
\hfill
\subfloat{\includegraphics[width=0.16\linewidth]{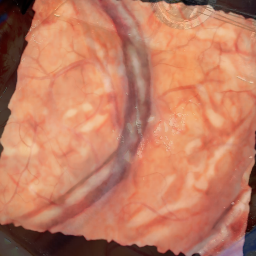}}
\hfill
\subfloat{\includegraphics[width=0.16\linewidth]{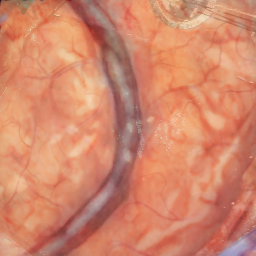}}
\hfill
\subfloat{\includegraphics[width=0.16\linewidth]{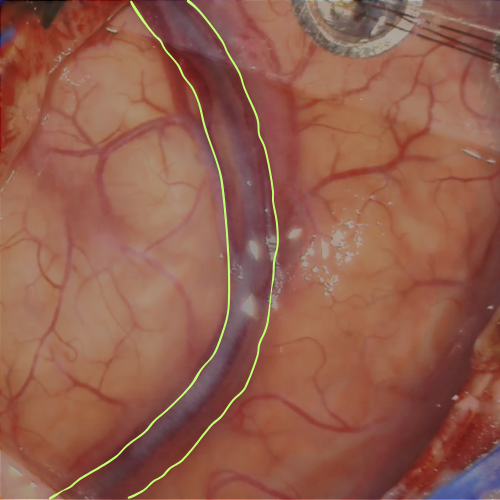}}\\
\hspace{-1.5em}
\subfloat{\includegraphics[width=0.16\linewidth]{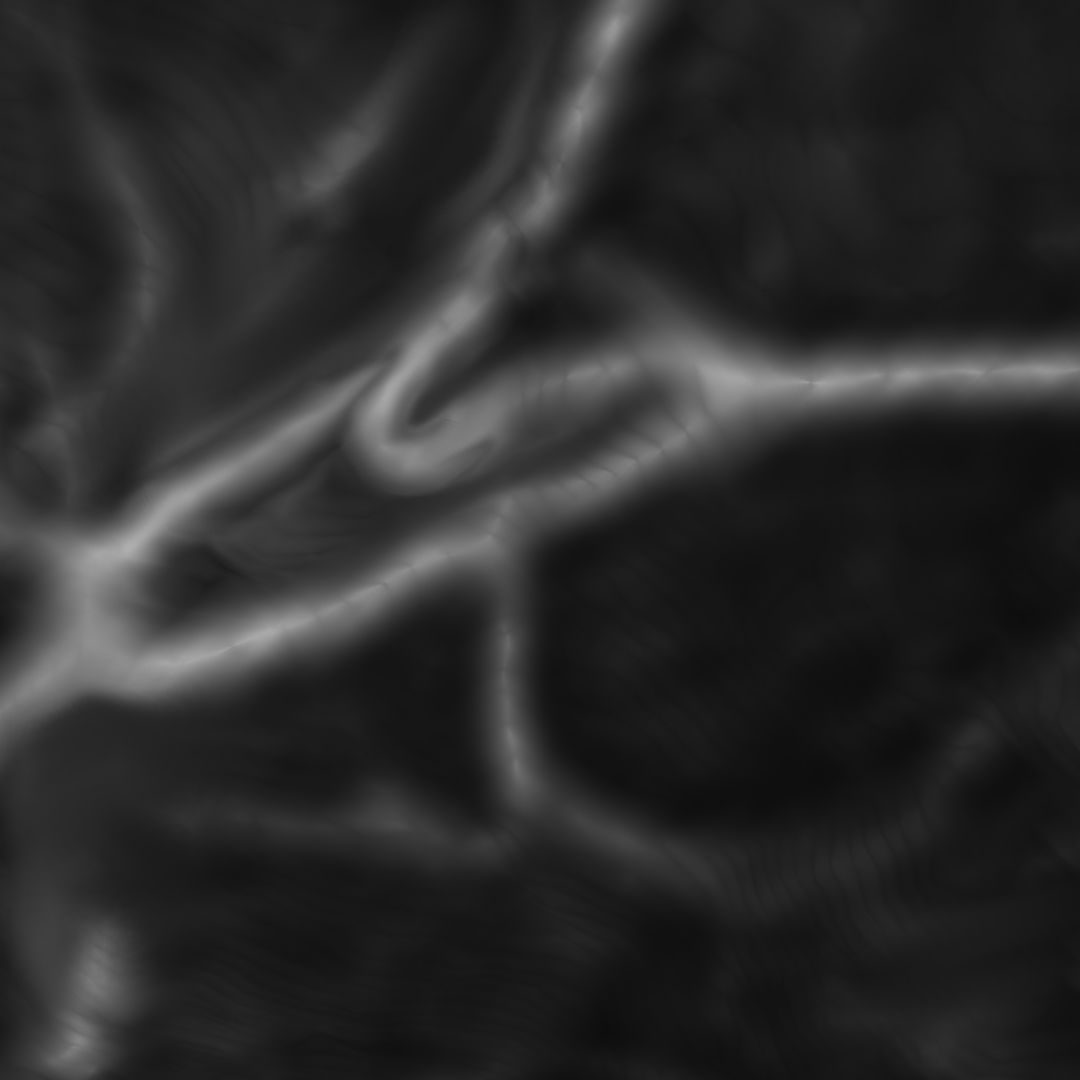}}
\hfill
\subfloat{\includegraphics[width=0.16\linewidth]{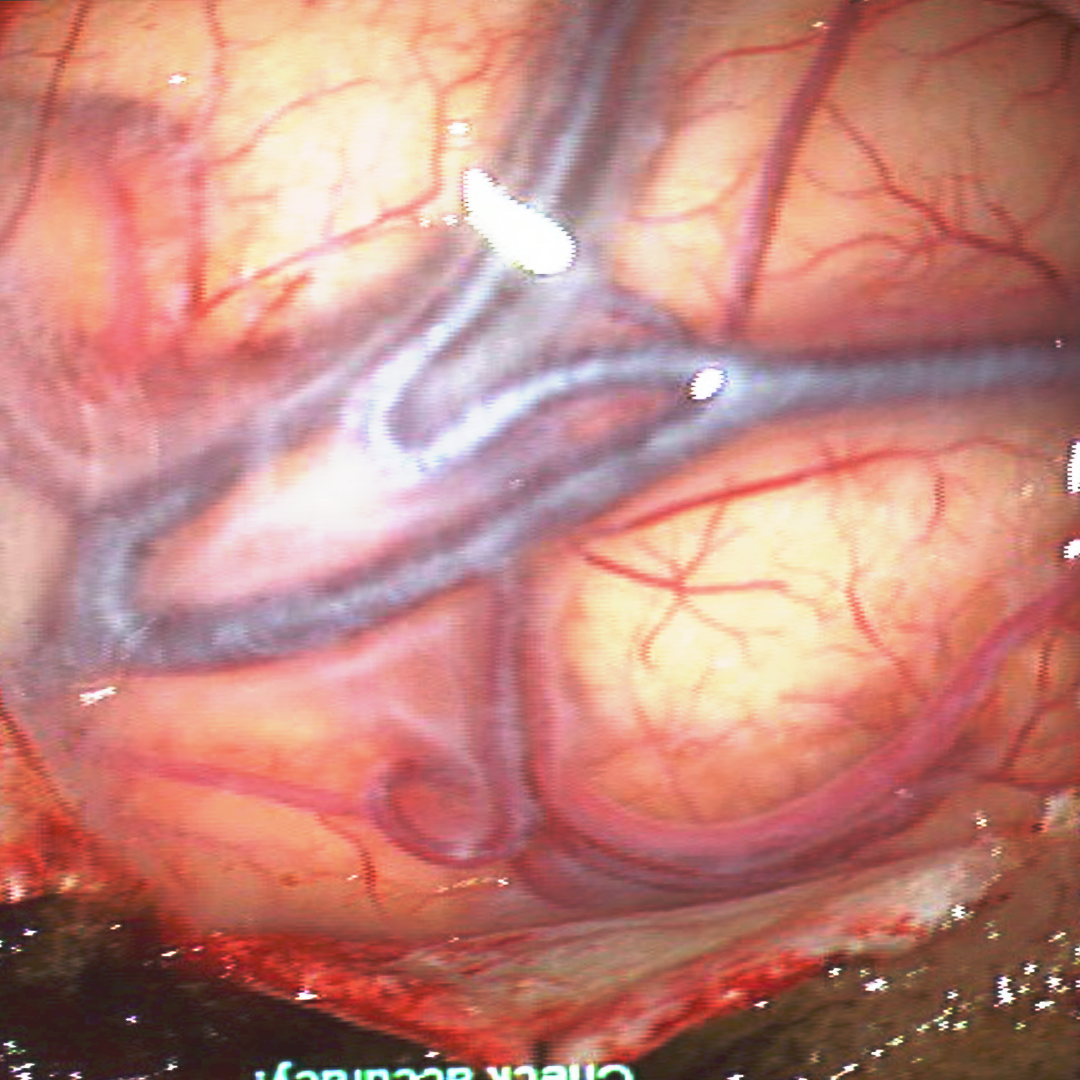}}
\hfill
\subfloat{\includegraphics[width=0.16\linewidth]{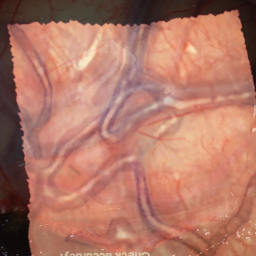}}
\hfill
\subfloat{\includegraphics[width=0.16\linewidth]{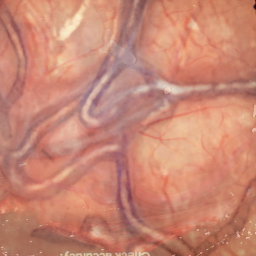}}
\hfill
\subfloat{\includegraphics[width=0.16\linewidth]{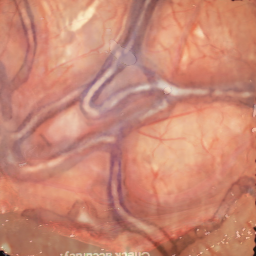}}
\hfill
\subfloat{\includegraphics[width=0.16\linewidth]{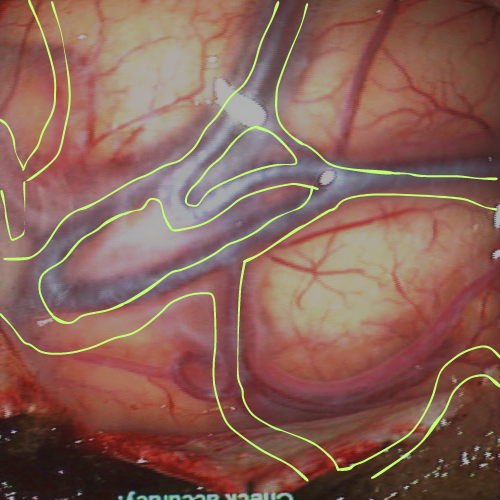}}\\
\subfloat{\includegraphics[width=0.16\linewidth]{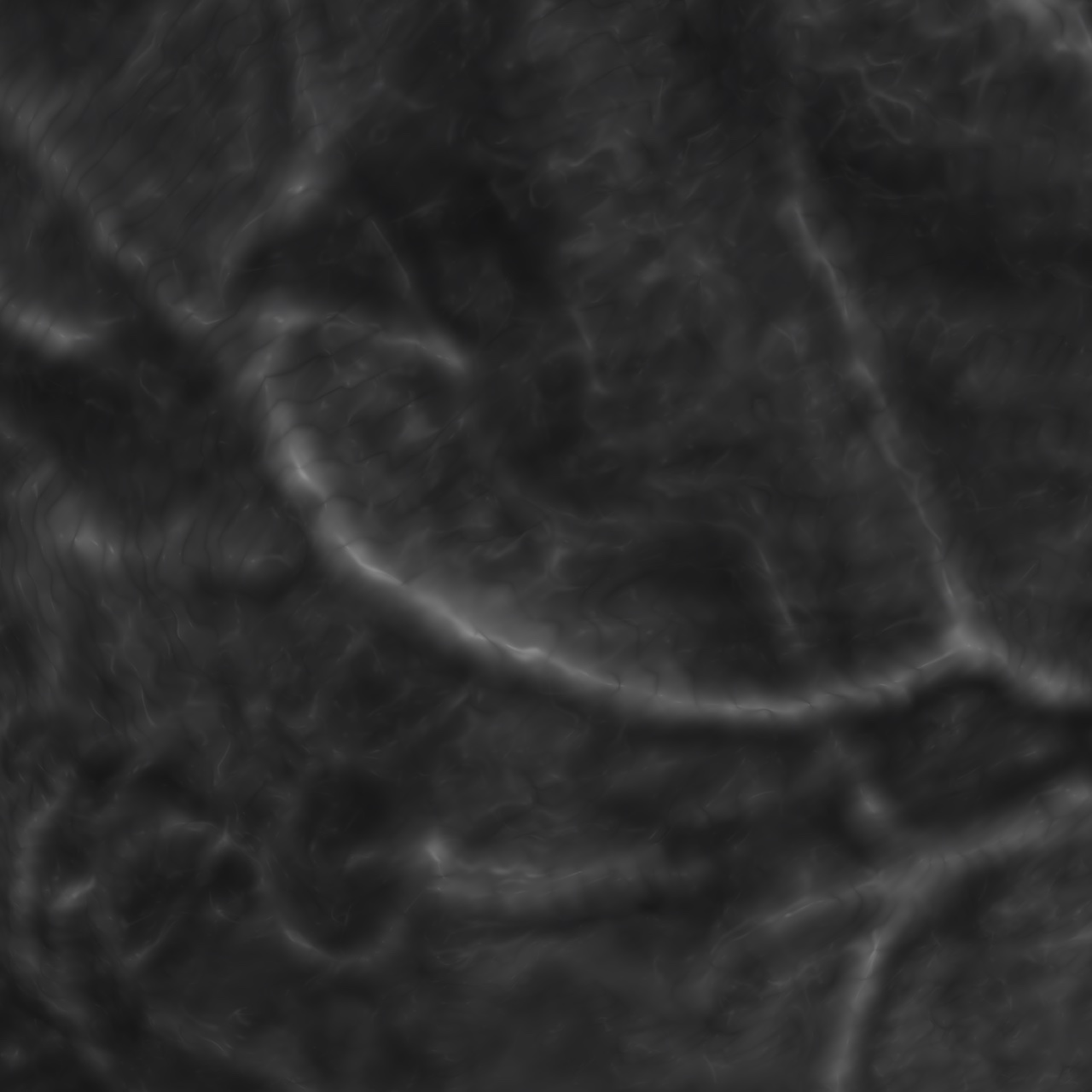}}
\hfill
\subfloat{\includegraphics[width=0.16\linewidth]{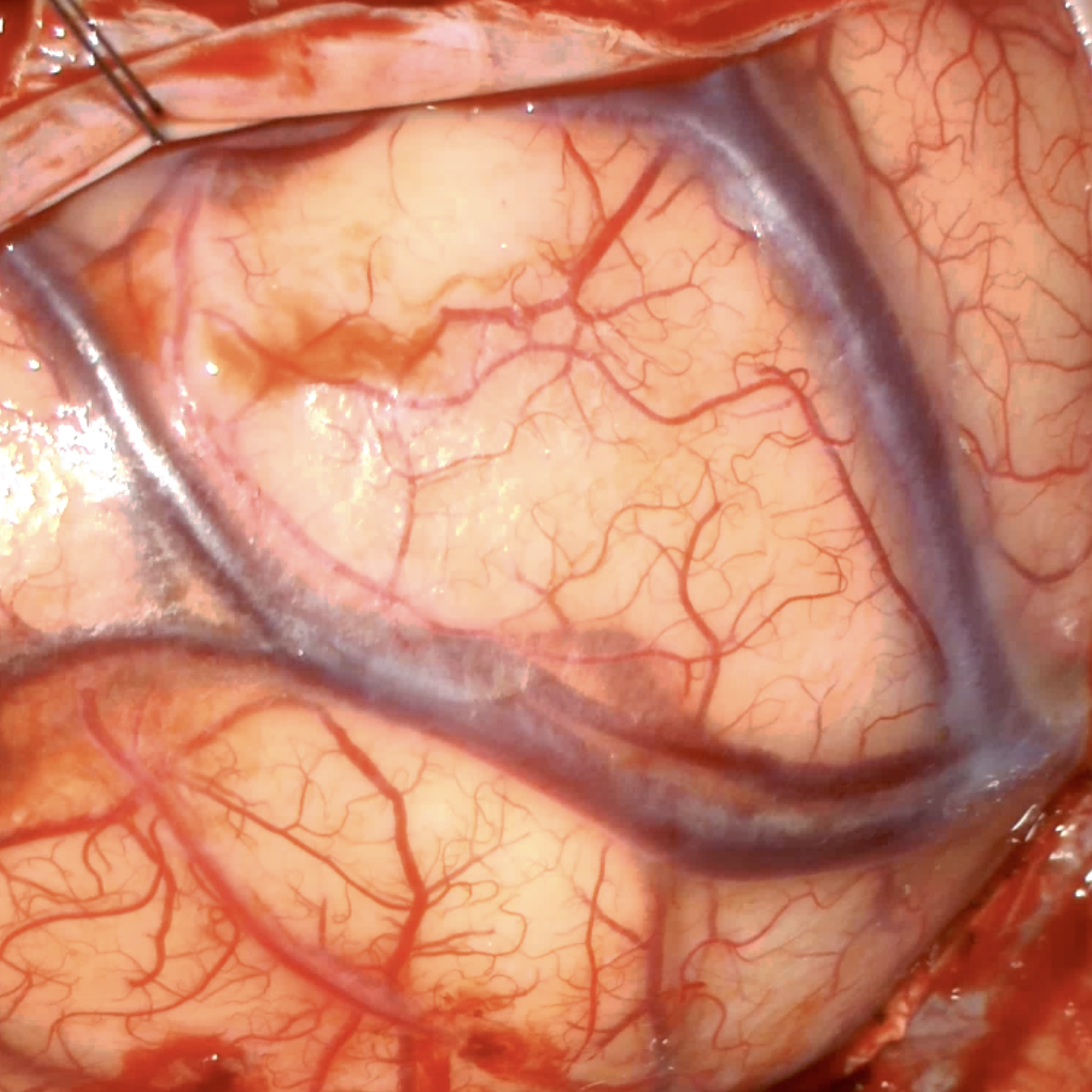}}
\hfill
\subfloat{\includegraphics[width=0.16\linewidth]{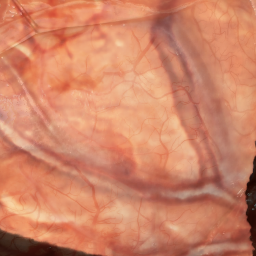}}
\hfill
\subfloat{\includegraphics[width=0.16\linewidth]{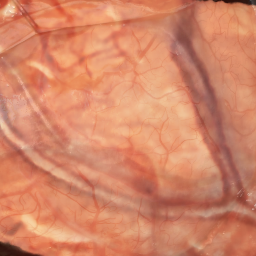}}
\hfill
\subfloat{\includegraphics[width=0.16\linewidth]{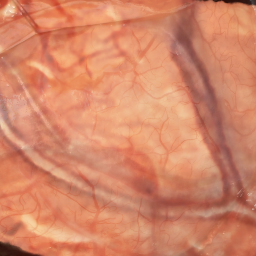}}
\hfill
\subfloat{\includegraphics[width=0.16\linewidth]{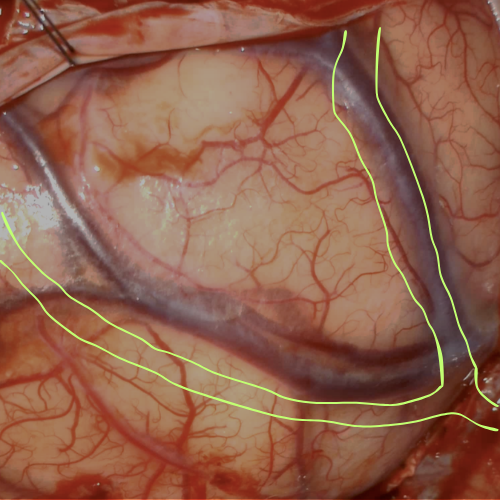}}\\
\hspace{-1.5em}
\subfloat{\includegraphics[width=0.16\linewidth]{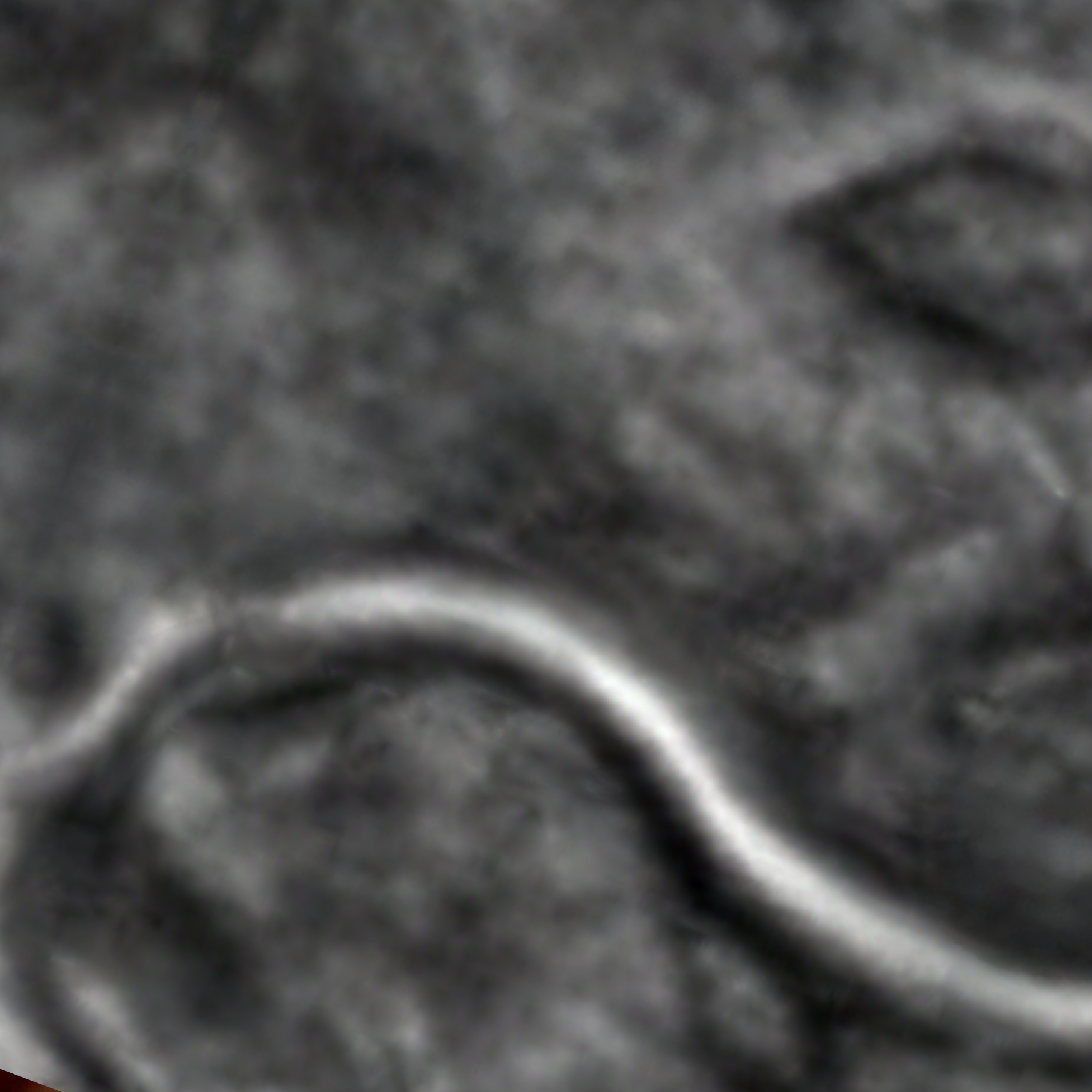}}
\hfill
\subfloat{\includegraphics[width=0.16\linewidth]{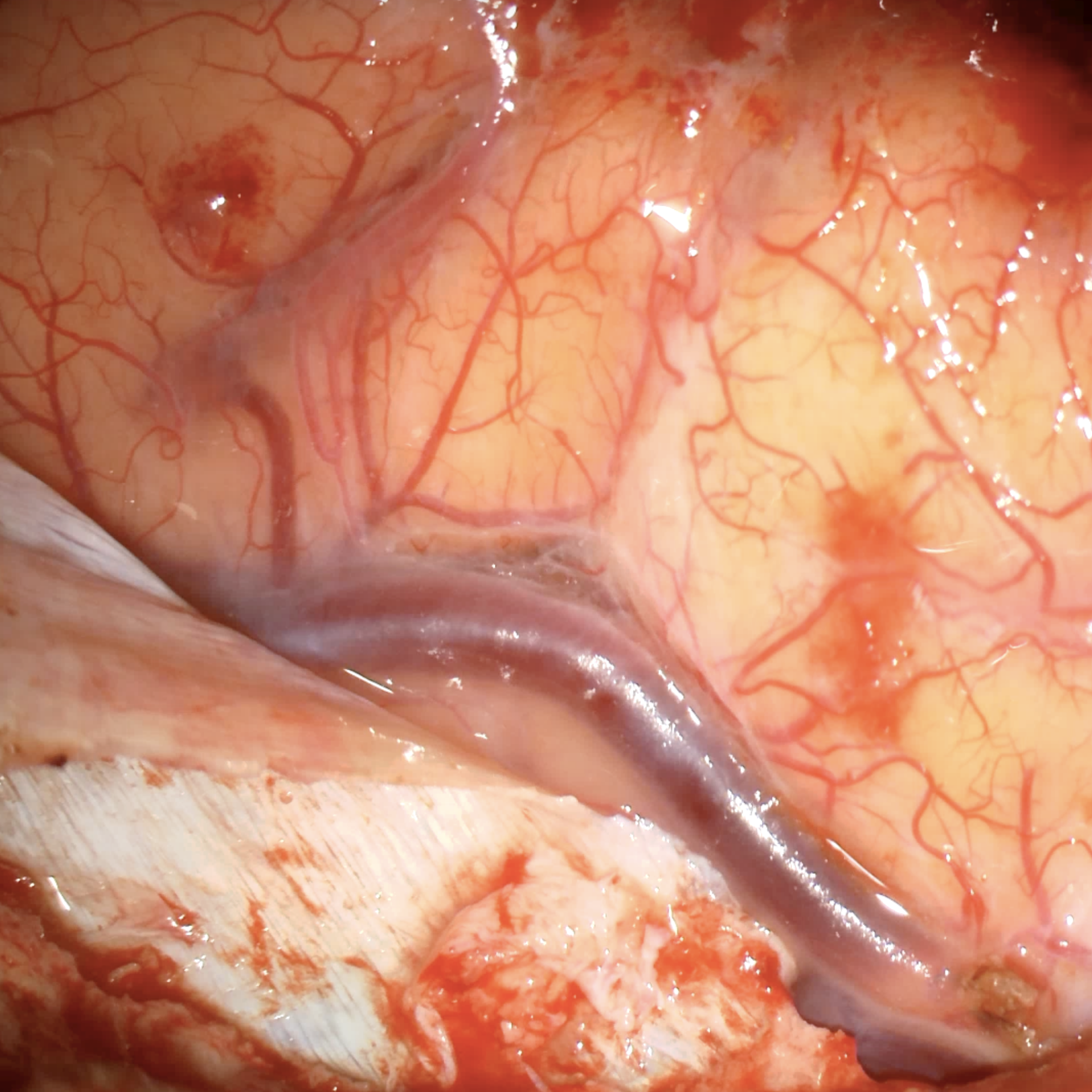}}
\hfill
\subfloat{\includegraphics[width=0.16\linewidth]{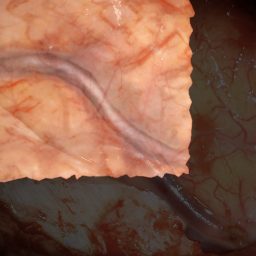}}
\hfill
\subfloat{\includegraphics[width=0.16\linewidth]{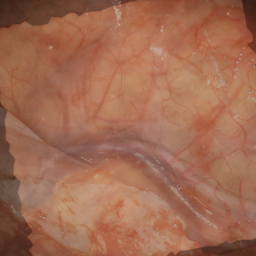}}
\hfill
\subfloat{\includegraphics[width=0.16\linewidth]{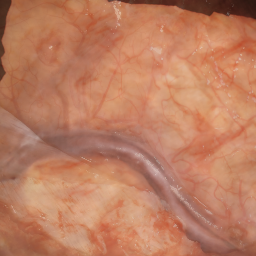}}
\hfill
\subfloat{\includegraphics[width=0.16\linewidth]{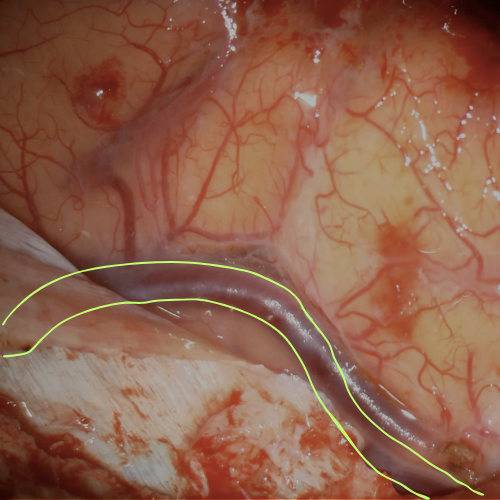}}\\
\hspace{-1.5em}
\caption{Tests on real cases, one case per row. From left to right: preoperative MR scan of the area (volume rendering), intraoperative target image from the surgical microscope, 3 optimization steps (early-optimization, mid-optimization, and final pose estimation), and intraoperative image with vessel-overlay of our estimated pose.}
\label{fig:clinical}
\end{figure}

We evaluate each case separately and exclude its style from the training set. 
The cases represent different craniotomy openings with varied appearances and anatomies.
We provide qualitative results with a visual assessment of each case in Fig. \ref{fig:clinical}.
Except for the case in row 3, all other pose estimations reach a visually correct registration. For case 3, we found that, depending on initialization, the optimization encountered a local minimum (fitting either the right vessel or the left). Since our method performs rigid registration, we hypothesize that this case might have undergone considerable deformation, requiring a non-rigid approach and therefore failing on a rigid one.

\captionsetup[subfigure]{labelformat=empty}

\section{Conclusion}
In this paper, we presented a novel 3D/2D intraoperative registration approach for neurosurgery by introducing a cross-modal
inverse neural rendering that disentangles NeRF representation into structure and appearance adapting thereby NeRFs to the preoperative and intraoperative settings. 
We presented experiments with qualitative and quantitative results on synthetic and retrospective real patient data, showing that our method outperforms the state-of-the-art, performs well in real conditions, and meets clinical needs.
Future work will extend this representation to handle deformation.
This can be achieved by modifying the density component of our implicit neural representation to be robust to non-rigid transformations.

\begin{credits}
\subsubsection*{Acknowledgments.}
This work was supported by the National Institutes of Health (R01EB032387, R01EB034223, R03EB033910, and K25EB035166).

\subsubsection*{Disclosure of Interests.}
The authors have no competing interests to declare that are
relevant to the content of this article.
\end{credits}

\bibliographystyle{splncs04}
\bibliography{mybib}

\end{document}